\def\year{2022}\relax
\documentclass[letterpaper]{article}
\usepackage{aaai22}
\usepackage{times}
\usepackage{helvet}
\usepackage{courier}
\usepackage[hyphens]{url}
\usepackage{graphicx} 
\usepackage{natbib}  
\usepackage{caption} 
\DeclareCaptionStyle{ruled}{labelfont=normalfont,labelsep=colon,strut=off} 
\usepackage{float}
\usepackage{comment}
\usepackage{epstopdf}
\usepackage{epsfig} 
\usepackage{subfigure}

\usepackage{algorithm}
\usepackage{algorithmic}

\usepackage{newfloat}
\usepackage{listings}
\floatstyle{ruled}
\newfloat{listing}{tb}{lst}{}
\floatname{listing}{Listing}

\usepackage{epsfig}
\usepackage[symbol]{footmisc}
\usepackage{amsmath}
\usepackage{subfigure}
\usepackage{amssymb}
\usepackage{algorithm}
\usepackage{times}
\usepackage{epsf}
\usepackage{epsfig}
\usepackage{epstopdf}
\usepackage{amsmath}
\usepackage{amssymb}
\usepackage{subfigure}
\usepackage{algorithm}
\usepackage{theorem}
\usepackage{multirow}
\usepackage{multicol}
\usepackage{enumerate}

\makeatletter

\newcommand{\IB}{\mathbf{I}}

\newcommand{\MB}{\mathbf{M}}

\newcommand{\SB}{\mathbf{S}}

\newcommand{\XB}{\mathbf{X}}

\newcommand{\aB}{\mathbf{a}}

\newcommand{\cB}{\mathbf{c}}

\newcommand{\pB}{\mathbf{p}}

\newcommand{\sB}{\mathbf{s}}
\newcommand{\tB}{\mathbf{t}}

\newcommand{\vB}{\mathbf{v}}

\newcommand{\xB}{\mathbf{x}}
\newcommand{\yB}{\mathbf{y}}
\newcommand{\zB}{\mathbf{z}}

\newcommand{\RBB}{\mathbb{R}}

\newcommand{\EBB}{\mathbb{E}}

\newcommand{\alphaB}{\mbox{\boldmath$\alpha$\unboldmath}}

\newcommand{\epsilonB}{\mbox{\boldmath$\epsilon$\unboldmath}}

\newcommand{\argmin}{\mathop{\rm argmin}}

\newcommand{\kl}{\mathrm{KL}}

\newcommand{\sparsemax}{\mathsf{sparsemax}}

\makeatother

\begin{document}

\title{Towards Controllable Agent in MOBA Games with Generative Modeling}

\author {
    Shubao Zhang
}
\affiliations{
    AI Lab, Tencent, Shenzhen, China\\
    bravemind@zju.edu.cn
}

\maketitle

\begin{abstract}
\begin{quote}  
We propose novel methods to develop action controllable agent that behaves like a human and has the ability to align with human players in Multiplayer Online Battle Arena (MOBA) games. By modeling the control problem as an action generation process, we devise a deep latent alignment neural network model for training agent, and a corresponding sampling algorithm for controlling an agent's action. Particularly, we propose deterministic and stochastic attention implementations of the core latent alignment model. Both simulated and online experiments in the game Honor of Kings demonstrate the efficacy of the proposed methods. 
\end{quote}
\end{abstract}

\section{Introduction}
Recent advances, especially in the field of artificial intelligence for games (a.k.a. Game AI), provide significant progress on the road towards artificial general intelligence. We have witnessed milestones of AI agents in playing various games, including Atari \cite{Mnihabc2013,Mnihabc2015}, Go \cite{Silverabc2016,Silverabc2018}, StarCraft \cite{Vinyalsabc2019}, Poker \cite{BrownSandholm2019}, Dota2 \cite{Bernerabc2019}, Honor of Kings \cite{Ye2abc2020}, etc. Nowadays, instead of achieving superhuman performance, research interests in the community start to focus on the problem of cooperation \cite{Dafoeabc2020,Dafoeabc2021}. Developing AI agents that cooperate with humans is beneficial to humanity. MOBA games such as Dota2, Legend of League and Honor of Kings,  are suitable testbeds for studying the cooperation problem, as they capture the complexity and continuous nature of the real world. 
 
Multi-agent cooperation is crucially important to master MOBA games. The standard game mode in MOBA 
is 5v5, i.e., a team of five players compete against another team of five players. To win, one team must destroy the base of another competitive team. It requires that each player not only has strong individual combat skills but also cooperate with other teammates very well to compete against the enemies. In a game, the players in one team can be AI agents, or humans, or a mixture of them, which leads to different cooperation paradigms. Typical AI systems like OpenAI Five \cite{Bernerabc2019} and Tencent JueWu-RL \cite{Ye2abc2020} have successfully solved the AI-AI cooperation problem with multi-agent reinforcement learning. However,  the human-AI cooperation (or collaboration) problem is still lacking of research focus, which is ubiquitous in real-world business application scenarios that AI agents are used to accompany humans to play game.

Human-AI cooperation leverages the interaction between human and AI agent to maximize joint utility. This type of AI agent is cooperative (or collaborative) AI, which emphasizes the interactive and cooperative perspective instead of the autonomous perspective. The research of cooperative AI  aims to build machine agents with the capabilities needed for cooperation and to foster cooperation in populations of (machine and/or human) agents. \citet{Crandallabc2018} developed AI agent that combines a reinforcement learning algorithm with mechanisms for generating and responding to signals conducive to human understanding. They showed that this agent can cooperate with humans and other machines  at levels that rival human-human cooperation in a variety of two-player repeated stochastic games. \citet{Carrollabc2019} indicated that incorporating human data or models into the training process is critical to improve the cooperation capability of AI agent when paired with humans. To verify this hypothesis, they built an environment based on the game Overcooked. Recently, \citet{Bardabc2020} set a new Hanabi challenge for AI research. They suggested the development of an AI agent with theory of mind reasoning. This ability of inferring the intentions of other agents is significantly important not only for success in Hanabi, but also in broader collaborative efforts, especially those with humans.  \citet{Puigabc2021} built VirtualHome-Social, a multi-agent household environment for training and testing AI agent with social perception (i.e., theory of mind reasoning) and the ability to collaborate with other agents. In common, all these works studied the cooperation paradigm that the human and machine agent have an equal relationship. In contrast, another line of research focuses on the AI alignment problem involving a principal-agent relationship, in which the human (principal) has normative priority over the machine (agent) \cite{Russell2019}. It aims to design AI agent acting in accordance with human intentions. The aligned AI relies heavily on goal alignment with human and mechanism of control. \citet{Everitt2018} gave a formal definition of alignment, studied the misalignment problems in current reinforcement learning agents systematically, and proposed methods to avoid goal misalignment. \citet{szlamabc2019} built an in-game AI assistant in Minecraft that can interact with players through natural language and do whatever players want it to do. \citet{Youngabc2020} developed human-like chess agent Maia with imitation learning.

In this paper, we study the human-AI cooperation problem in MOBA games. A principled approach for cooperation is learning with the self-play methods such as multi-agent reinforcement learning. By virtue of some specific training strategies like population-based training, the self-play methods extended nicely to collaboration: human-AI teams performed well in Dota2 \cite{Bernerabc2019} and Capture the Flag \cite{Jaderbergabc2019}. However, in these games, the advantage mainly comes from AI agents' individual ability rather than coordination with humans. Additionally,  AI agents treat the human partners to be optimal or similar with themselves, which may result in weak cooperation with humans. A way to improve cooperation performance is using human-like agent trained by imitation learning, which performs better than agent trained by self-play \cite{Carrollabc2019}. The self-play or human-like agent may not always act in accordance with human since they have an equal relationship. It is thus desirable to develop agent that makes strong cooperation with humans, augmenting their capabilities.

To this end, we develop action controllable agent belonging to aligned AI. Human can control agent's action through command signals representing their intentions. We creatively combine imitation learning with the latent alignment model to attain human-like agent with the ability to align with human. More specifically, the main contributions of this paper are as follows:
\begin{itemize}
\item We model the agent control problem as an action generation process, leveraging the hidden selective attention mechanism in playing MOBA games. 
\item We design a deep latent alignment neural network model to learn the mechanism of action generation. 
\item We devise deterministic and stochastic attention implementations of the core latent alignment module 
that functions as feature selection and fusion. 
\item We propose an adaptive selective attention sampling algorithm serving as mechanism of control.  
\end{itemize}
Both simulated and online experiments demonstrate that deterministic and stochastic agents are able to execute human's commands well. Moreover, we show that for the first time, agent with stochastic action generation can play MOBA games well. The video demo of our agents executing human's commands can be assessed in https://sourl.cn/fKXYwa.

\section{Background}
\textbf{Notation.~~} The $i$-th entry of a vector ${\vB}$ is denoted by $v_i$. The $i$-th row of a matrix ${\MB}$ is denoted by $\MB_{i.}$. $||\vB||_0=\sum_{i=1}^n \IB_{v_i\neq0}$ denotes the $\ell_0$ norm of a vector $\vB \in \RBB^n$, where $\IB$ is the indicator function. 
Let $\kl(p(\xB)||q(\xB))=\sum_{i} p(x_i)\log \frac{p(x_i)}{q(x_i)}$. 
Let $\bigtriangleup^{n-1} := \{\vB \in \RBB^n :  \textbf{1}^T\vB=1, \vB\ge \textbf{0} \}$ be the $(n-1)$-dimensional probability simplex. Let $\sparsemax(\zB) := \argmin_{\pB\in\bigtriangleup^{n-1}} ||\pB-\zB||_2^2$ for $\zB\in\RBB^n$ \cite{MartinsAstudillo2016}. 


\subsection{Latent Alignment}
\begin{figure}[hpt]
\centering
\includegraphics[width=4.2cm]{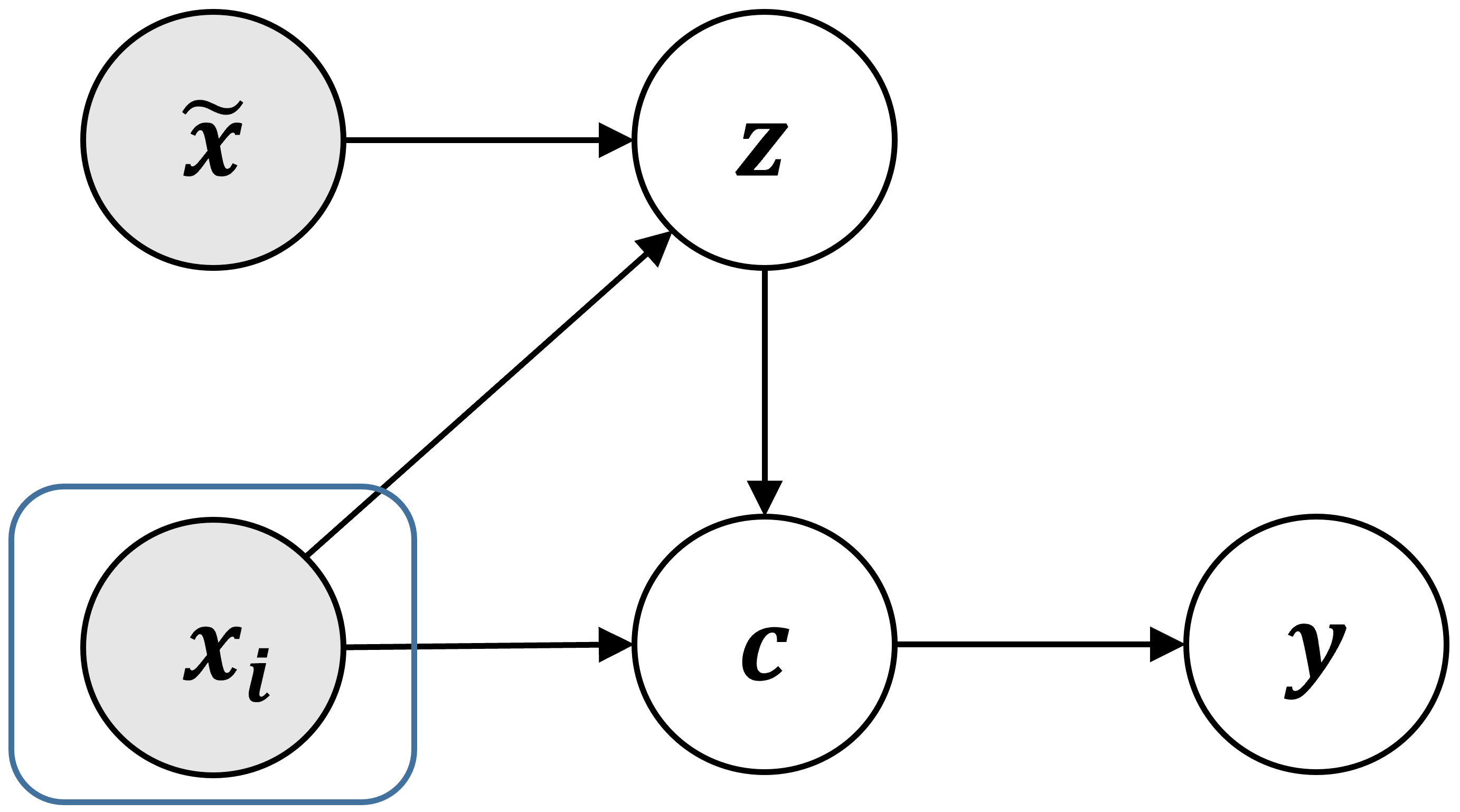} 
\vskip -0.05in
\caption{Latent alignment model.} 
\label{fig:latent_alignment_model}
\vskip -0.08in
\end{figure}

Figure \ref{fig:latent_alignment_model} shows a latent alignment model \cite{Dengabc2018}. Its mathematical form is expressed as follows:
\begin{equation}
\label{equation:latent_alignment_model}
\zB \sim \mathcal{D}(a(\xB, \tilde{\xB}; \theta)), \quad \cB = g(\xB,\zB; \theta), \quad \yB \sim f(\cB;\theta).
\end{equation}
Here, $\xB := \{\xB_1,\cdots,\xB_n\}$ $(\xB_i\in\RBB^d)$ is the observed input, $\tilde{\xB}\in\RBB^d$ is an arbitrary ``query'', $\zB\in\bigtriangleup^{n-1}$ is the latent alignment variable indicating which member (or mixture of members) of $\xB$ generates the context vector $\cB\in\RBB^m$, and $\yB\in\mathcal{Y}$ is the predicted output. The function $a$ produces the parameters for an alignment distribution $\mathcal{D}$. The alignment function $g$ calculates the context vector given the set of input vectors and the alignment variable. The function $f$ gives a distribution over the output.
\vskip -0.08in
 Given the training samples $\{(\xB,\tilde{\xB},\hat{\yB})\}$, the model parameters $\theta$ can be obtained by maximizing the log marginal likelihood:
\begin{equation}
\label{lam_obj}
\max_{\theta} \log p(\yB=\hat{\yB} | \xB, \tilde{\xB} ) =  \max_{\theta} \log \EBB_{\zB} [ f( g(\xB, \zB))_{\hat{\yB}} ].
\end{equation}

\subsection{Stochastic Variational Inference}
A principled approach to optimize the latent variable model (\ref{lam_obj}) is to use amortized variational inference \cite{KingmaWelling2014,Rezendeabc2014}. It optimizes the evidence lower bound (ELBO):
\begin{equation}
\begin{aligned}
\label{elbo_b}
& \log  \EBB_{\zB \sim p(\zB | \xB,\tilde{\xB}) }  [ p(\yB | g( \xB, \zB) ) ] \\
& \geq \EBB_{\zB \sim q(\zB) } [ \log p(\yB | g( \xB, \zB) ) ] - \kl [ q(\zB) || p(\zB | \xB,\tilde{\xB}) ].
\end{aligned}
\end{equation}
The bound is tight when the variational distribution $q(\zB) \in \mathcal{Q}$ (with the constraint $\textrm{supp } q(\zB) \subseteq \textrm{supp } p(\zB | \xB, \tilde{\xB}, \yB)$) is equal to the true posterior $p(\zB | \xB, \tilde{\xB}, \yB) \in \mathcal{D}$. Amortized variational inference uses an inference network to produce the parameters of the variational distribution $q(\zB ; \alphaB)$ with $\alphaB = encoder(\xB, \tilde{\xB}; \phi)$. Let $\yB = predictor(g( \xB, \zB); \psi)$. To tighten the gap of (\ref{elbo_b}), it optimizes the objective:
\begin{equation}
\label{obj1_b}
\max_{\phi,\psi} \EBB_{\zB \sim q(\zB;\alphaB) } [ \log p_{\psi}(\yB | g( \xB, \zB) ) ] - \kl [ q_{\phi}(\zB;\alphaB) || p (\zB | \xB,\tilde{\xB}) ] .
\end{equation}

\section{Method}   
In this section, we first introduce the selective attention mechanism in playing MOBA games which is the motivation of our methods. Next, we present the deep latent neural network which utilizes a latent alignment model to align unit features with the predicted action. Then, we discuss implementations of the latent alignment model. Finally, we propose a novel latent variable sampling strategy for manipulating the predicted action.

\begin{figure*}[th]
\centering
\includegraphics[width=16.6cm]{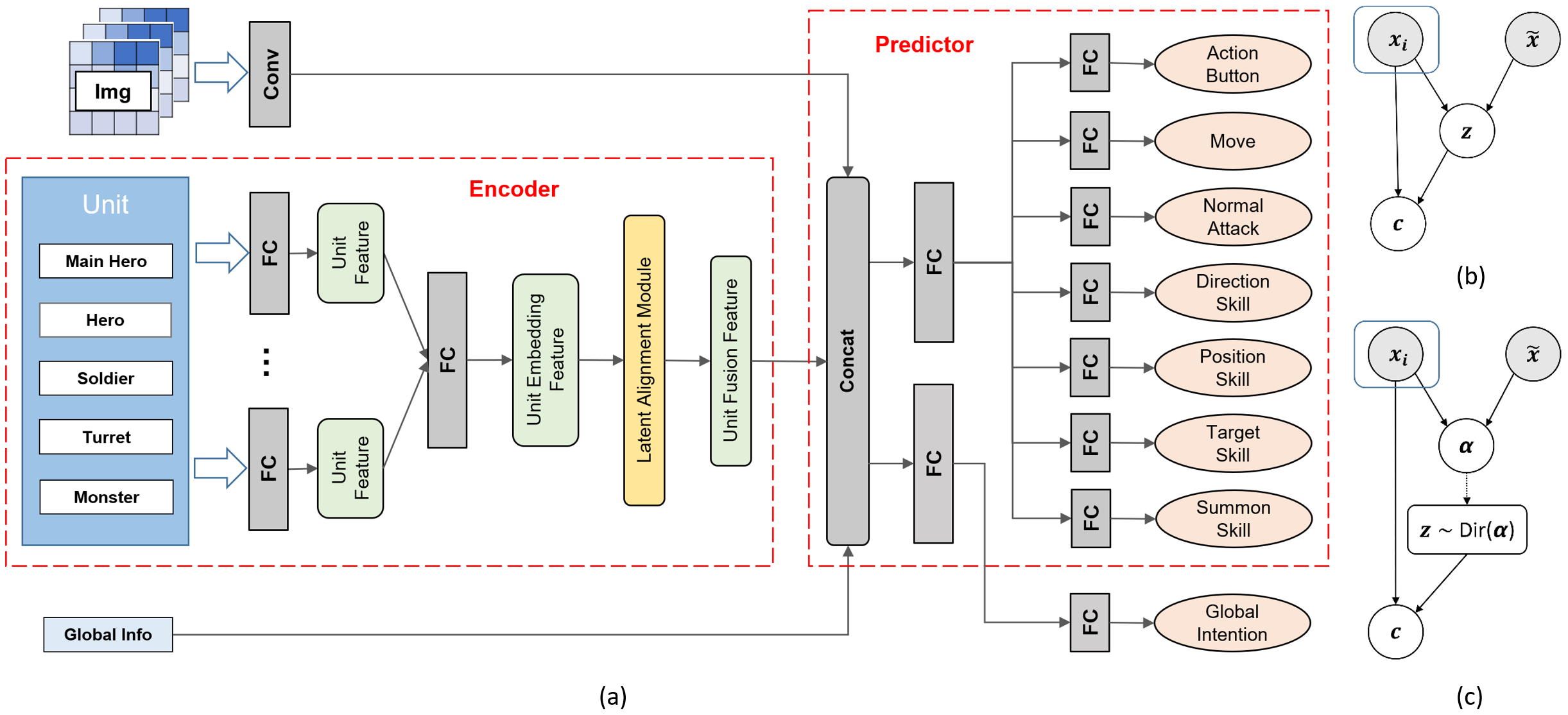} 
\vskip -0.1in
\caption{(a) Deep latent alignment neural network architecture (``Conv'' indicates the convolutional layers, and ``FC'' indicates the fully-connected layers). The left red dashed box indicates the Encoder, and the yellow block in it indicates the latent alignment module.The right red dashed box indicates the Predictor. (b) Latent alignment module with deterministic attention. (c) Latent alignment module with stochastic attention (``Dirichlet'' is abbreviated as ``Dir'').} 
\label{fig:dnn_latent}
\vskip -0.15in
\end{figure*}

\subsection{Selective Attention Mechanism}
Our goal is to train human-aware agent that obeys human's command. That is, agent does what human wants it to do. To this end, we need to dig out the explanatory factors that determine agent's action. We can thus manipulate the predicted action through changing the key factors. 

Observe a phenomenon that when playing MOBA games, the player usually allocates attention to a limited set of important units (including heros, soldiers, organs and monsters) 
in the game, and makes a decision depending on those specific units. This selective attention mechanism, i.e., the player only perceives an important part of all units when making decisions, is the first principle of playing MOBA games. Inspired by it, we propose a deep latent neural network architecture and a corresponding latent variable sampling strategy in the next subsections.

\subsection{Deep Latent Alignment Neural Network}
Let $D := \{(\sB_i,\aB_i)\}_{1\dots N}$ be a set of state-action samples. The task is to learn a policy function $\pi_{\theta}(\aB_t|\sB_t)$ modeled by a deep neural network with parameters $\theta$ on the dataset $D$. The policy network $\pi_{\theta}$ takes observation state $\sB_t$ as input, and predicts a probability distribution over action $\aB_t$ as output.

\textbf{State and Action Space.~~} We adopt the design of features and labels in Tencent JueWu-SL \cite{Ye1abc2020}, since the aim of this work is to attain human-aware agent that follows player's command in addition to be capable of playing MOBA games. The observation states $\{\sB_i\}_{1,\dots,N}$ are represented by multi-modal features, which consists of vector feature and image-like feature. The vector features are made up of units attributes and game states. The image-like features are made up of several channels including main hero's local view map, hero's skill damage area and obstacles. The action labels $\{\aB_i\}_{1,\dots,N}$ have a hierarchical multi-class structure to control a hero's behavior. Specifically, each label consists of two sublabels representing level-1 and level-2 actions. The level-1 action indicates which action to take, including move, normal attack, skill 1, skill 2, skill 3, summon skill, and so on. The level-2 action tells how to execute action concretely according to its action type. If the first sublabel is skill 1 (or 2, or 3), the second one will depend on the action type of level-1 action because all the skills 1-3 of different heros belong to: direction type, target type, and position type. For example, if it is the position type, the second one will be a position where the action will take effect.

\textbf{Network Architecture.~~}  A schematic view of the deep latent alignment neural network $\pi_{\theta}$ is shown in Figure \ref{fig:dnn_latent}, which adopts an encoder-decoder structure. The encoder produces the encoded context vector given the unit embedding features as input. The predictor gives the distribution over the output conditioned on the context vector, image-like features and global info. Aiming to manipulate the predicted action of $\pi_{\theta}$, a latent alignment module is inserted into the last layer of the encoder.

\subsection{Latent Alignment Model with Attention}
We discuss two alternative implementations for the latent alignment module: deterministic attention (Figure \ref{fig:dnn_latent}b) and stochastic attention (Figure \ref{fig:dnn_latent}c) .

\textbf{Deterministic Attention.~~} A straightforward implementation of the selective attention mechanism is to set the latent alignment module as a linear weighted feature fusion layer:
\begin{equation}
\label{deterministic_attention}
 \cB_{fusion}=\sum_{i=1}^n z_{unit_i} *  \xB_{unit_i} 
\end{equation}
subject to
\begin{align}
\label{probability_simplex}
 &\zB=[z_{unit_1},\dots,z_{unit_n}] \in \{  \vB : \vB \in \bigtriangleup^{n-1},  ||\vB||_0 \le k \}.
\end{align}
Here, $\cB_{fusion}\in\RBB^d$ denotes the fused feature, $\xB_{unit_i}\in\RBB^d$ denotes the \textit{i}-th unit embedding feature, and $\zB\in\RBB^n$ is a weight vector with each element corresponding to a specific unit. The alignment variable $\zB$ is not treated as a latent random variable, but calculated deterministically according to: 
\begin{equation}
\label{cal_latentvariable}
\zB = \sparsemax(\tB)  \textrm{ with } t_i = s(\xB_i,\tilde{\xB}), i=1,\ldots,n
\end{equation}
where $s$ is a similarity function, and $\tilde{\xB}$ takes the main hero's embedding feature. For the choices of similarity measures, refer to \cite{Chaudhariabc2021}. 
Equation (\ref{deterministic_attention}) is actually soft attention \cite{Bahdanauabc2015}, which is an approximation of alignment \cite{Kelvinabc2015,Dengabc2018}. 

As the alignment variable $\zB$ is not a random variable, the objective has only the policy loss term:
\begin{equation}
\label{obj_deterministic}
\begin{aligned}
 \max_{\phi,\psi}  \mathcal{L}(\phi, \psi; \xB, \tilde{\xB}, \yB) :=    \log p_{\phi,\psi} (\yB=\hat{\yB} | \xB, \tilde{\xB} )  
\end{aligned}
\end{equation}
This model is smooth and differentiable, and can be trained end-to-end by using standard back-propagation. For more details of the policy loss, refer to \cite{Ye1abc2020}.

\textbf{Stochastic Attention.~~} It is natural to assume that the alignment variable $\zB$ obeys a sparse Dirichlet distribution, according to the constraint (\ref{probability_simplex}). A stochastic implementation of the latent alignment module is thus as follows:
\begin{equation}
\label{stocastic_attention}
 \cB_{fusion}=\sum_{i=1}^n z_{unit_i} *  \xB_{unit_i}, \quad  \zB \sim Dirichlet(\alphaB)
\end{equation}
with $\alphaB \geq \textbf{0}$ and $0 < ||\alphaB||_0 \leq k$. 
The distribution parameter $\alphaB$ is calculated according to the way in (\ref{cal_latentvariable}). 

To obtain the parameters of the deep latent alignment neural network with stochastic attention, we need to optimize the ELBO (\ref{obj1_b}). Take both $\mathcal{Q}$ and $\mathcal{D}$ as Dirichlets, and assume the true posterior $p (\zB | \xB,\tilde{\xB},\yB) \sim Dir(\alphaB^0)$. It is intractable to directly optimize (\ref{obj1_b}) with variational inference, since the Dirichlet distribution $q(\zB;\alphaB)$ does not exist a differentiable non-centered reparameterization function. To tackle this problem, we resort to rejection sampling variational inference that uses the proposal function of a rejection sampler as the reparameterization function \cite{Naessethabc2017}.


The Dirichlet distribution can be simulated using gamma random variables since if $z_i \sim \Gamma(\alpha_i, 1)$ i.i.d. then $\zB = [\frac{z_{1}}{\sum_{i} z_i},\dots,\frac{z_{n}}{\sum_{i} z_i}] \sim Dir(\alphaB)$. An efficient rejection sampler for the Gamma distribution $\Gamma(\alpha,1)$ with $\alpha\geq 1$ is:
\begin{equation}
\begin{aligned}
\label{gamma_rs}
 z =& h_{\Gamma}(\epsilon,\alpha) := (\alpha - \frac{1}{3}) ( 1 + \frac{\epsilon}{\sqrt{9\alpha - 3}})^3,  \\
& \epsilon \sim s(\epsilon) := N(0,1).
\end{aligned}
\end{equation}
However, sampling with $h_{\Gamma}(\epsilon,\alpha)$ is not equivalent to sampling with the Gamma distribution since some samples are rejected. The distribution of an accepted sample $\epsilon \sim  s(\epsilon)$ is obtained by marginalizing over the uniform variable $u$ of the rejection sampler, 
\begin{equation}
\label{accepted_distribution}
\pi (\epsilon;\alpha) = \int \pi (\epsilon, u;\alpha) du = s(\epsilon) \frac{q( h_{\Gamma}(\epsilon, \alpha) ; \alpha )}{ M_{\alpha} r( h_{\Gamma}(\epsilon, \alpha) ; \alpha)} ,
\end{equation}
where $r(z; \alpha)$ is the proposal distribution for the rejection sampler, $z=h_{\Gamma}(\epsilon, \alpha), \epsilon \sim s(\epsilon)$ is the reparameterization of $r(z; \alpha)$, and $M_{\alpha}$ is a constant such that $q(z; \alpha)\leq M_{\alpha} r(z; \alpha)$.  

The ELBO in (\ref{obj1_b}) can be rewritten as the expectation of the transformed variable $\epsilonB$: 
\begin{equation}
\label{obj2}
\begin{aligned}
 \max_{\phi,\psi}  \mathcal{L}(\phi, \psi; \xB, \tilde{\xB}, \yB) :=  & \EBB_{\epsilonB \sim  \pi(\epsilonB;\alphaB) } [ \log p_{\psi} (\yB | g( \xB, h(\epsilonB;\alphaB) ) ] \\
&  - \kl [ q_{\phi} (\zB;\alphaB) || p (\zB | \xB,\tilde{\xB}) ] .
\end{aligned}
\end{equation}
The gradient of the $predictor$ network is given as:
\begin{equation}
\begin{aligned}
\label{predictor_gradient}
&\nabla_{\psi}  \mathcal{L}(\phi, \psi; \xB, \tilde{\xB}, \yB) = \nabla_{\psi}   \EBB_{\zB \sim  q(\zB;\alphaB) } [ \log p_{\psi} (\yB | g( \xB, \zB)) ] .
\end{aligned}
\end{equation}
There is no need to reparameterize this term as the samples used to estimate $\mathcal{L}$ depend on $q$ which is parameterized by $\phi$. The gradient of the $encoder$ network is given as:
\begin{equation}
\begin{aligned}
\label{encoder_gradient}
&\nabla_{\phi} \mathcal{L}(\phi, \psi;  \xB, \tilde{\xB}, \yB)  =  - \nabla_{\phi} \kl [ q_{\phi} (\zB;\alphaB) || p (\zB | \xB,\tilde{\xB}) ] \\
& \quad\quad\quad + \nabla_{\phi} \EBB_{\epsilonB \sim  \pi(\epsilonB;\alphaB) } [ \log p_{\psi} (\yB | g( \xB, h(\epsilonB;\alphaB) ) ].  \\ 
\end{aligned}
\end{equation}
As shown in \cite{Naessethabc2017}, the second term in (\ref{encoder_gradient}) can be decomposed as: 
\begin{equation}
\begin{aligned}
\nabla_{\phi} \EBB_{\epsilonB \sim  \pi(\epsilonB;\alphaB) } [ \log p_{\psi} (\yB | g( \xB, h(\epsilonB;\alphaB) ) ]  =  \mathcal{G}_{\textrm{rep}}^{\phi} + \mathcal{G}_{\textrm{cor}}^{\phi},
\end{aligned}
\end{equation}
where $\mathcal{G}_{\textrm{rep}}$ and $\mathcal{G}_{\textrm{cor}}$ for the case of a one-sample Monte Carlo estimator are given as:
\begin{align}
& \mathcal{G}_{\textrm{rep}}^{\psi} =  \nabla_{\zB} \log p_{\psi} (\yB |g(\xB, \zB)) \nabla_{\phi} h(\epsilonB, \alphaB),  \\
& \mathcal{G}_{\textrm{cor}}^{\psi} =  \log p_{\psi} (\yB | g(\xB, \zB) ) \nabla_{\phi} \log \frac{q(h(\epsilonB, \alphaB))}{r(h(\epsilonB, \alphaB))}.
\end{align}
$\mathcal{G}_{\textrm{rep}}$ corresponds to the gradient assuming that the proposal is exact and always accepted, and $\mathcal{G}_{\textrm{cor}}$ corresponds to a correction part of the gradient that accounts for not using an exact proposal. 
The KL-divergence term has analytical form:
\begin{equation}
\begin{aligned}
\label{kl_div}
& \kl [ q_{\phi} (\zB;\alphaB) || p (\zB | \xB,\tilde{\xB}) ] = \log(\Gamma(\sum_{i} \alpha_i)) \\ 
& - \log(\Gamma(\sum_{i} \alpha_i^0))  + \sum_{i} \log\Gamma (\alpha_i^0)  - \sum_{i} \log\Gamma (\alpha_i)  \\
&  + \sum_i ( \alpha_i - \alpha_i^0) (\Psi(\alpha_i) - \Psi(\sum_i \alpha_i)),
\end{aligned}
\end{equation}
where $\Psi$ is the digamma function.

A trick to improve the efficiency of rejection sampler is to use shape augmentation for the Gamma distribution \cite{Naessethabc2017}. A $\Gamma(\alpha, 1)$ distributed variable $z$ can be expressed as $z = \tilde{z} \prod_{i=1}^B u_i^{\frac{1}{\alpha + i -1}}$ for a positive integer $B$, $u_i \overset{i.i.d.}{\sim} U[0,1]$, $\tilde{z} \sim \Gamma(\alpha + B, 1)$. To sampling $z \sim \Gamma(\alpha, 1)$ with $\alpha > 0$, we use the rejection sampler for $\Gamma(\alpha + B, 1)$ instead. With increasing parameter $\alpha+B$, the acceptance rate of rejection sampler increases, and the correction term $\mathcal{G}_{\textrm{cor}}$ decreases that leads to a lower variance gradient.

\begin{figure}[hpt]
\centering
\includegraphics[width=3.5cm]{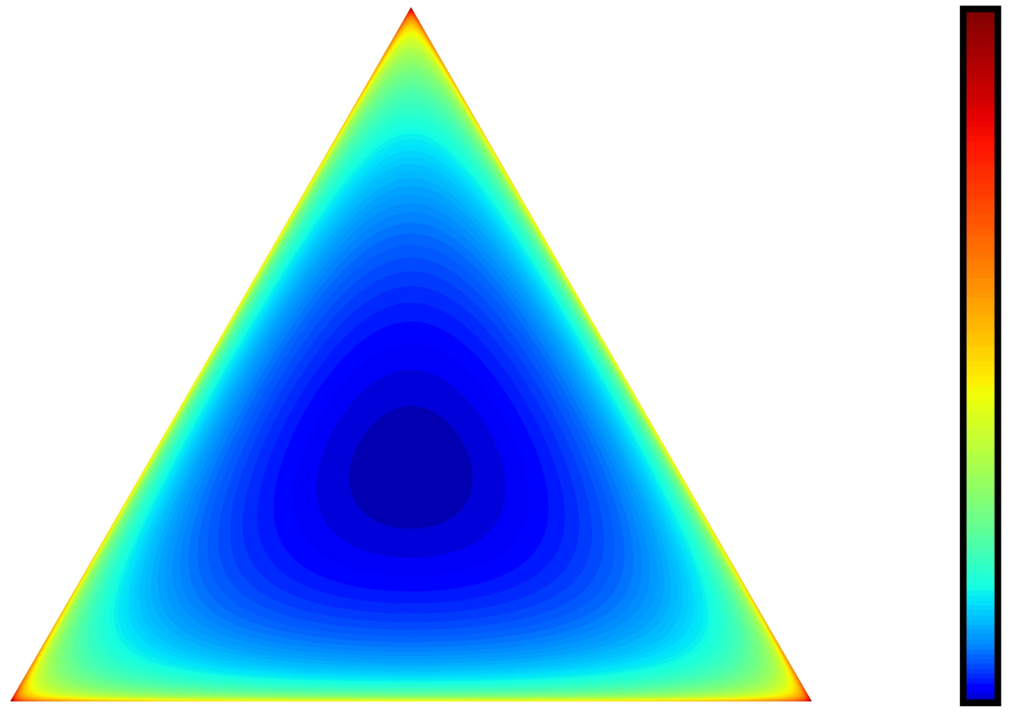} 
\vskip -0.05in
\caption{The probability mass of $Dirichlet(0.1, 0.1, 0.1)$ over the $2$-dimensional probability simplex.} 
\label{fig:probability_simplex}
\vskip -0.08in
\end{figure}

\textbf{Discussion.~~} The constraint set (\ref{probability_simplex}) of the alignment variable $\zB$ constitutes a latent space, which requires that $\zB$ must be sparse and in the probability simplex. This specific latent space has an intriguing structure. As shown in Figure \ref{fig:probability_simplex}, the probability mass is concentrated at the vertices of the probability simplex.
Consequently, the sparse latent alignment variables $\zB$ of all state-action $(\sB_i,\aB_i)$ pairs related to a unit are clustered in a corresponding reduced probability simplex. 
The clustering property provides the potential to change the predicted action by sampling a related latent variable $\zB$ from the latent space and using it to predict. 
In addition, sparsity of $\zB$ ensures the predicted action interpretable, as only the relevant unit features attributed to the action are kept and the irrelevant ones are masked out with zero coefficients through the feature fusion process. 
\vskip -0.08in

It is worth pointing out that different choices of the alignment function $g$ lead to different attention types: soft (or additive) attention \cite{Bahdanauabc2015}, multiplicative attention \cite{Luongabc2015}, scale (or channel) attention \cite{Huabc2018}, etc.

\subsection{Mechanism of Control}
In Honor of Kings, command signals issued by the player are categorized into three types: attack, retreat, assemble. A valid command signal must indicate a target unit (a hero, or soldier, or organ, or monster) in the game. The execution of a command is a process that agent moves towards the target unit. Therefore, in each command execution frame, it is necessary to ensure that the agent makes an action depending on the target unit. As the latent alignment variables of all state-action pairs related to a specific unit are clustered in a reduced probability simplex, it is feasible to retrieve an action working on the target unit by sampling a proper latent alignment variable and using it to predict. 

Inspired by the selective attention mechanism, we design a novel latent variable sampling algorithm shown in Algorithm \ref{alg:alpa}\footnote{* indicates that this line is optional.}. It consists of two steps. First, obtain a set of units that agent should attend. In order to make agent moving towards the target unit, it is ok to only add the target unit into the attention set. However, to let agent better handling with other units nearby, it needs to add units nearby both main hero and the target unit into the attention set such that agent can perceive them. This strategy makes agent behaving more human-like, and less likely to act abnormally as it has higher probability of sampling a proper latent variable. Second, sampling a latent alignment variable that is used to generate desired action. For the deterministic attention method, we get a sparse attention weight by first setting attention scores of units not in the attention set into a enough small value and then applying $\sparsemax$ operator to the modified attention scores\footnote{The attention scores are calculated in the way of Transformer \cite{Ashishabc2017}. Other computing methods are optional.}. For the stochastic attention method, we first get a sparse Dirichlet distribution parameter corresponding to the target unit's reduced probability simplex, and then sampling a sparse latent alignment variable from the Dirichlet distribution with the modified sparse parameter. In practice, to reduce uncertainty, we sampling multiple samples and average them as the final output.

\textbf{Command Execution Rule.~~}We use the global intention of JueWu-SL to judge whether agent should execute received commands. We define the judgement rule as: agent execute a command only if the target unit's position is in the set of top-k positions predicted by the global intention auxiliary task shown in Figure \ref{fig:dnn_latent}.

\begin{algorithm}[t] 
   \caption{Adaptive Selective Attention Sampling}
   \label{alg:alpa}
\begin{small}
\begin{algorithmic}[1]
   \STATE {\bfseries Input: ~} $\tB = {[\XB\XB^T]_{1.}}/{\sqrt{d}}$, $\XB=[\xB_{unit_1};\dots;\xB_{unit_n}]$.
     \STATE {//\textbf{Step 1. }\textrm{Obtain the selective attention set $\SB_{\textrm{attention}}$}} 
     \STATE  $\SB_{\textrm{attention}} = \{ \textrm{unit}_{\textrm{mainhero}}, \textrm{unit}_{\textrm{target}} \}$
     \FOR{ $i=1,\cdots,n$ } 
     \STATE \textrm{//Add units nearby main hero into $\SB_{\textrm{attention}}$}
     \IF{ dist( $\textrm{unit}_i, \textrm{unit}_{\textrm{mainhero}} ) < \textrm{radius}$ }
     \STATE \textrm{Add $\textrm{unit}_i$ into $\SB_{\textrm{attention}}$}.
  	\ENDIF
	\STATE \textrm{//Add units nearby the target unit into $\SB_{\textrm{attention}}$}
	\IF{ dist( $\textrm{unit}_i, \textrm{unit}_{\textrm{target}} ) < \textrm{radius}$ }
     \STATE \textrm{Add $\textrm{unit}_i$ into $\SB_{\textrm{attention}}$}.
  	\ENDIF
    \ENDFOR
     %
     \STATE {//\textbf{Step 2. }\textrm{Sampling a sparse latent variable $\zB_{\textrm{sampled}}$}}
     \FOR{ $i=1,\cdots,n$ } 
     \IF{ $\textrm{unit}_i \notin \SB_{\textrm{attention}}$ }
     \STATE $t_i = -\infty$
     \ENDIF
    \ENDFOR    
    \STATE \textrm{*Deterministic Attention: } $\zB_{\textrm{sampled}} = \sparsemax(\tB)$
    \STATE \textrm{*Stochastic Attention: } $\alphaB = \sparsemax(\tB)$ \\ 
    ~~~~~~~~~~~~~~~~~~~~~~~~~~~~~~~~~~~~~~$\zB_{\textrm{sampled}} \sim Dirichlet(\alphaB)$
    %
   \STATE {\bfseries Output: ~} $\zB_{\textrm{sampled}}$.
\end{algorithmic}
\end{small}
\end{algorithm}

\section{Related Work}
The most related to our work is Tencent JueWu-SL, which provides a supervised learning method to achieve human-level performance in playing Honor of Kings. As our goal is to improve human-agent cooperation, 
it is suitable to use JueWu-SL's method to train human-like agents. Therefore, we directly use the same design of features, labels, training data preprocessing and the policy loss. For details of these parts, refer to \cite{Ye1abc2020}. Although JueWu-SL agents perform collaboration with humans well, their actions can not be controlled by human players. In other words, JueWu-SL agents make decisions autonomously, whose actions only depend on the observed game states. In contrast, our proposed methods can train action controllable agent that can do what a human player wants it to do. Technically, the main difference with JueWu-SL is that our method is based on the latent alignment model. Consequently, we can change the agent's predicted action by manipulating the latent alignment variable. 

Conditional imitation learning is a promising approach to train an action controllable agent \cite{Felipeabc2018}. 
The predicted action is determined by both the state and a conditional command. 
Human can control agent's action via a command. This method requires that the training samples must be triples $\{($state, action, command$)\}$. At training time, the model is given not only the state and action, but also a command representing human's intention. In comparison, our methods need no command infos for training. The command is used to sampling a latent variable when inferring.

The stochastic attention method enjoys the same spirit with latent dirichlet allocation in topic modeling \cite{Bleiabc2003}. Given the observed state, the action is affected by a mixture of a small number of important units. Recently, variational autoencoder with the Dirichlet distribution as prior has been widely studied. One major difficulty is reparameterization trick of the Dirichlet distribution. Various approaches are proposed to tackling this issue, including softmax Laplace approximation \cite{SrivastavaSutton2017}, Weibull distribution approximation \cite{Zhangabc2018}, implicit reparameterization gradients \cite{Figurnovabc2018}, inverse Gamma approximation \cite{Jooabc2020}, etc. We use the rejection sampling variational inference \cite{Naessethabc2017}, as which has been validated to performing well \cite{BurkhardtKramer2020}.

\section{Experiments}
\textbf{Experimental Setup.~~} The training datasets is made in the same way with JueWu-SL. For each of all heros, about 10 million samples are extracted from games played by the top human players and with players' overall performance score above threshold. All heros are classified into five categories: warrior hero, shooter hero, wizard hero, assistant hero, jungle hero. Accordingly, the samples of heros belonging to the same category are put together, shuffled, and stored. The unified model for each of five categories is trained by using the optimizer Lamb \cite{Youabc2020}, with a initial learning rate $0.0002$ and a batch size $128$ for each of eight P40 GPU cards. When training the stochastic model, the encoder of well-trained deterministic model is provided as the prior.

\textbf{Model Evaluation.~~}
We take JueWu-SL agents as baseline, which are deployed online to provide services to hundreds of millions of players in Honor of Kings. Particularly, our agents are compared with JueWu-SL-H, JueWu-SL-M and JueWu-SL-L, as whose levels are almost matched with high, medium and low human-level, respectively. For both our and JueWu-SL agents, unified models for five categories are trained on the same datasets. To be fair, the number of parameters of the deterministic and stochastic 
models are about the same with JueWu-SL-H. 

Table \ref{tab:validation_acc} shows that our methods are comparable with JueWu-SL in learning human's policy. Both deterministic and stochastic methods fit human's policy worse than JueWu-SL-H, since the feature fusion process of linear weighted sum losses more information than of concat. In addition, the stochastic method is more difficult to optimize, as it involves the approximation of objective and distribution. 
\begin{table}[hpt]\setlength{\tabcolsep}{1.6pt}
\caption{Validation accuracy of jungle hero unified model.} 
\vskip -0.16in
\label{tab:validation_acc}
\begin{center}
\begin{small}
\begin{tabular}{lccccccr}
\hline
 & Action & Move & Attack &  SkillDir  &  SkillPos & SkillTar  \\
\hline
Deterministic &  0.83  &  0.277  &  0.978  & 0.19 & 0.136	& 0.883 \\
Stochastic   &  0.8142   &  0.2567  &  0.9706  & 0.1681	& 0.1246	& 0.8385 \\
JueWu-SL-H &  \textbf{0.834}  &  \textbf{0.284}  &  \textbf{0.978}  &  \textbf{0.196}  &  \textbf{0.147}	& \textbf{0.91} \\
JueWu-SL-M &  0.825  &  0.272  &  0.978  & 0.18	& 0.135  & 0.872 \\
JueWu-SL-L  &  0.811  &  0.253  &  0.961  & 0.165 & 0.127	& 0.833 \\
\hline
\end{tabular}
\end{small}
\end{center}
\vskip -0.13in
\end{table}

\begin{figure*}[th]
\centering
  \subfigure{
    \label{fig:determinstic_attention}
    \begin{minipage}{0.5\textwidth}
\centering
      \includegraphics[width=6cm]{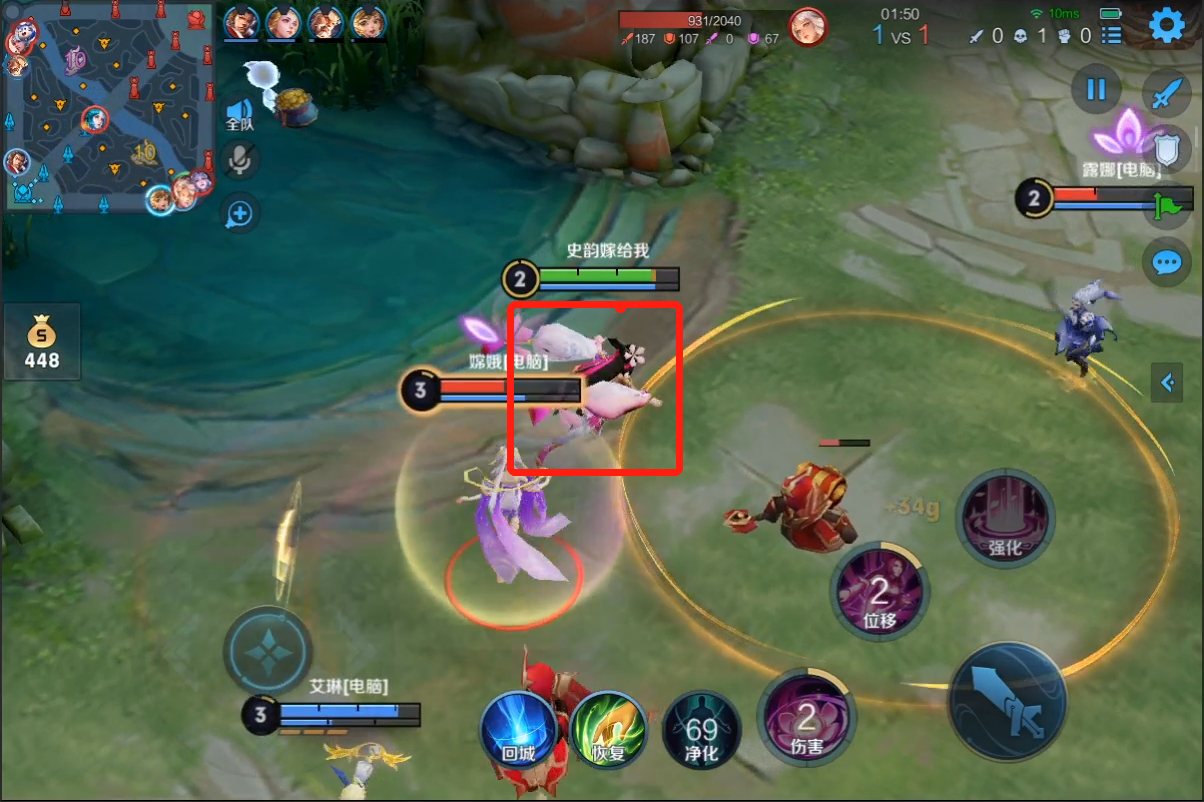}
      \center{\small
		\begin{tabular}{lccccr}
			\hline
     			Unit &   $z_{unit}$  & Sum of $z_{unit}$ \\
			\hline
			EnemyHero1  &  0.4718  & 0.4718 \\
			EnemyHero2  & 0.2342  & 0.7061 \\
			EnemySoldier1  & 0.1454  & 0.8515 \\
			Monster1  & 0.0312 & 0.8827 \\
			EnemyOrgan1  & 0.0151 & 0.8978 \\
			EnemyBase  & 0.0138 & $\textbf{0.9116}\geq0.9$ \\
			FriendHero1 & 0.0124 & 0.924 \\
			EnemySoldier2 & 0.0097 & 0.9337 \\
			MainHero & 0.0089 & 0.9426 \\
			EnemyBlueBuff & 0.0042 & 0.9468 \\
			\hline
		\end{tabular}
     }
     \center{\small(a) Deterministic}
    \end{minipage}}%
  \subfigure{
    \label{fig:stochastic_attention}
    \begin{minipage}{0.5\textwidth}
\centering
      \includegraphics[width=6cm]{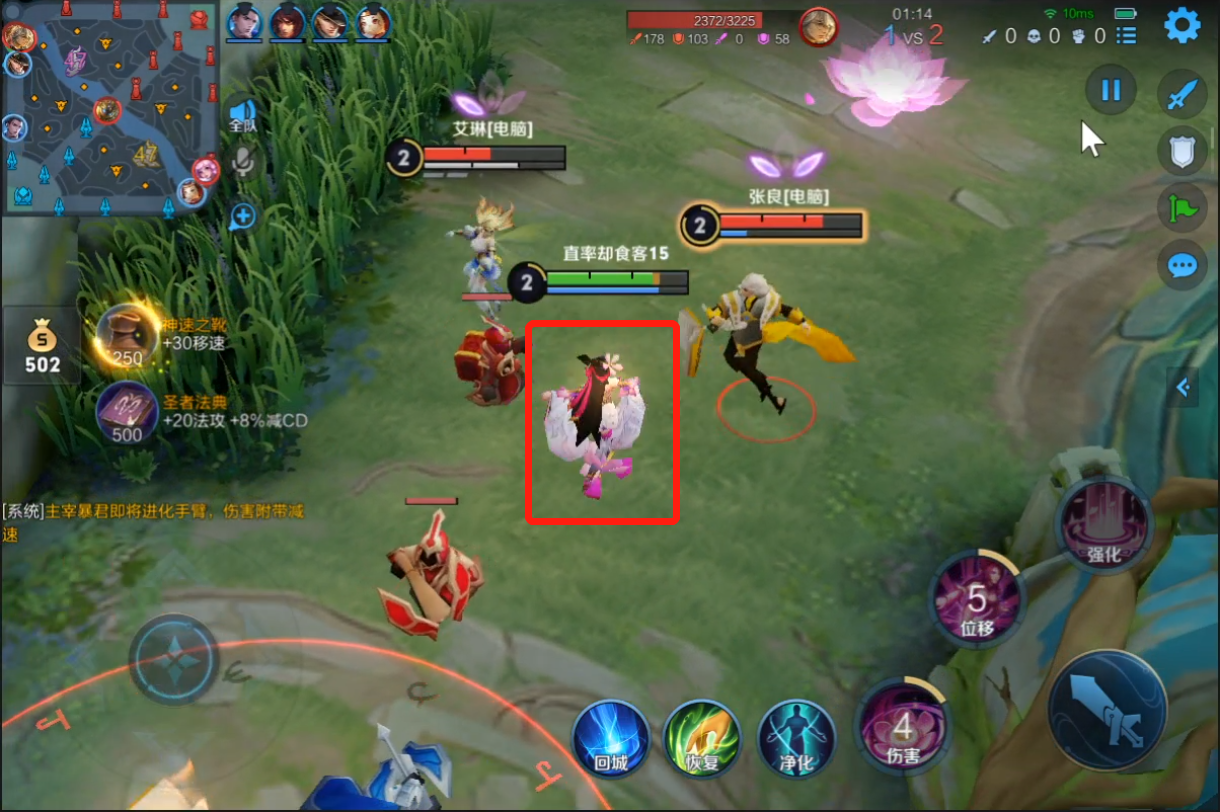}
      \center{\small
		\begin{tabular}{lccccr}
			\hline
     			Unit &  $\alpha_{unit}$  &  $z_{unit}$  & Sum of $z_{unit}$ \\
			\hline
			EnemyHero1  &   0.4279  & 0.4325  & 0.4325\\
			EnemySoldier1 & 0.1856 & 0.2016  & 0.634 \\
			EnemyHero2  & 0.1557 & 0.1441  & 0.7781 \\
			FriendHero1  & 0.0493 & 0.0421 &  0.82 \\
			Monster1 & 0.0267 & 0.0269 & 0.8471 \\
			FriendBase & 0.0195 & 0.0251 & 0.8722 \\
			FriendOrgan1 & 0.0164 & 0.1424 & 0.8865 \\
			EnemySoldier2 & 0.0087 & 0.0136 & $\textbf{0.9}\geq0.9$ \\
			FriendOrgan2 & 0.0055 & 0.0099 & 0.91 \\
			Monster2 & 0.0057 & 0.0097 & 0.9197  \\
			\hline
		\end{tabular}
     }
     \center{\small(b) Stochastic}
    \end{minipage}}%
\vskip -0.15in
  \caption{Unit inference by the deterministic and stochastic methods, respectively. The hero in red box is controlled by AI agent. The table below the image lists the top-10 important units (119 units in total) that agent selects to attend. 
  $z_{unit}$ denotes the attention weight corresponding to a unit. $\alpha_{unit}$ denotes the parameter of the Dirichlet distribution  corresponding to a unit.}
  \label{fig:unit_inference} 
  \vskip -0.15in
\end{figure*}

Table \ref{tab:combat} validates the efficacy of our agents in playing Honor of Kings. 
Although randomness is introduced into policy prediction, 
stochastic agents act with little uncertainty at most of time. 
However, in states with  ambiguity, their actions may be uncertain. 
For example, agents probably wander around a local area in the navigation states. 
\vskip -0.08in

\begin{table}[hpt]\setlength{\tabcolsep}{1.6pt}
\caption{Win rate of our agents against JueWu-SL agents.}
\vskip -0.16in
\label{tab:combat}
\begin{center}
\begin{small}
\begin{tabular}{lccccr}
\hline
     &  Win Rate  \\
\hline
Deterministic \textit{V.S.} JueWu-SL-H  &   0.008  \\
Deterministic \textit{V.S.} JueWu-SL-M  &   0.616   \\
Deterministic \textit{V.S.} JueWu-SL-L  &   0.995  \\
\hline
Stochastic \textit{V.S.} JueWu-SL-H  &  0   \\
Stochastic \textit{V.S.} JueWu-SL-M  &   0.103  \\
Stochastic \textit{V.S.} JueWu-SL-L  &   0.527  \\
\hline
Deterministic \textit{V.S.} Stochastic   &   0.991  \\
\hline
\end{tabular}
\end{small}
\end{center}
\vskip -0.25in
\end{table}

Figure \ref{fig:unit_inference} verifies that both deterministic and stochastic methods learn the selection attention mechanism.  
The latent alignment variable $\zB$ is sparse
, as the sum of top-10 units' coefficients dominates the rest. 
Agent's action heavily depends on units with high weight coefficients.

\textbf{Command Response and Execution Performance.~~} 
We define the following indices: 
\vskip -0.18in
\begin{align*}
& \textrm{Response rate} = \frac{ \textrm{\#commands excuted in total} } {\textrm{\#commands received in total}}, \\
& \textrm{Success rate} = \frac{\textrm{\#commands executed successfully in total} } {\textrm{\#commands executed in total} }, \\
& \textrm{Abnormal rate} = \frac{ \textrm{\#commands executed abnormally in total} } { \textrm{\#commands executed in total} }. \\
\end{align*}
\vskip -0.2in
\noindent A command is executed successfully only if agent reaches the target unit within the maximum execution time. 
A command is executed abnormally if agent fails to reach the target unit within the maximum execution time because of that it acts abnormally, e.g., lingering. 
Abnormal command execution is due to that improper latent variables are sampled to generate abnormal actions. 
Commands neither executed successfully nor abnormally belong to those ending normally. For example, during command execution, the maximum execution time may run out if agents meet with enemy units. 

We make a simulated experiment: 1.simulated commands are generated according to some scripted rules; 2.agents execute all received commands. Table \ref{tab:command_execution_simulated} shows that the deterministic method outperform the stochastic method, as stochastic agents often act abnormally.
\vskip -0.08in

\begin{table}[hpt]\setlength{\tabcolsep}{1.6pt}
\caption{Simulated command execution performance.}
\vskip -0.16in
\label{tab:command_execution_simulated}
\begin{center}
\begin{small}
\begin{tabular}{lccccr}
\hline
Jungle hero  & Indices &  Assemble  &  Retreat & Attack  \\
\hline
Deterministic & Success rate  &  \textbf{0.684}   &  \textbf{0.995}  & \textbf{0.761}  \\
& Abnormal rate  &       \textbf{0.072}  &  \textbf{0}  &  \textbf{0.055}   \\
\hline
Stochastic  & Success rate  &  0.382  &  0.458  &  0.406   \\
& Abnormal rate  &      0.242  &  0.194  &  0.21   \\
\hline
\end{tabular}
\end{small}
\end{center}
\vskip -0.13in
\end{table}

Since the standards for product to be online are very strict, only deterministic agents are qualified to be deployed online to test. 
In the online gameplay scenarios, two human players are teamed up with three AI agents to compete against five JueWu-SL agents.
Online deterministic agents execute commands conditioned on the output of command execution rule. 
They can reject unreasonable commands and abort command being executed. 
Table \ref{tab:command_response_online} shows online command response performance.
Table \ref{tab:command_execute_online} illustrates command execution performance. 
Note that online Success rate is lower than the simulated ones, since commands being executed may be aborted by the command execution rule. 
In the real experiences, the game designers think that our agents respond to commands like real humans.

\textbf{Cooperation Performance.~~}
We make an online ABTest experiment: two groups of players use our and JueWu-SL agents as their teammates, respectively. 
Table \ref{tab:cooperation_performance} shows that our collaborative agents make positive effect on the performance of players and the team, as helping to improve win rate and shorten the game time.
\vskip -0.08in

\begin{table}[H]\setlength{\tabcolsep}{1.6pt}
\caption{Cooperation Performance.}
\vskip -0.15in
\label{tab:cooperation_performance}
\begin{center}
\begin{small}
\begin{tabular}{lccccccr}
\hline
 &  Win Rate &  Frames &  Kill  & Dead & Assist &  Gold  \\
\hline
Team  & \textbf{0.767}  &  \textbf{12545}   &  \textbf{32.8}  &  22.4 & \textbf{32.3}  &  \textbf{59407}  \\
\textbf{Deterministic} Agents &   &   & 9.9  & 10.9  & 19.7  & 27263.4 \\
Human Players  &   &   & 22.9 & 11.5  & 12.6  & 32143.6 \\
\hline
Team  &  \textbf{0.755}  &  13709 &  31.3  &  \textbf{20.2}  & 28.7  & 54856.9  \\
\textbf{JueWu-SL} Agents &   &   & 9.5  & 8.1  & 17  & 23772.7 \\
Human Players  &   &   & 21.8 & 12.1  & 11.7  & 31084.2 \\
\hline
\end{tabular}
\end{small}
\end{center}
\vskip -0.1in
\end{table}

\section{Conclusion}
In this paper, we propose methods to develop action controllable agent. 
Leveraging the hidden selective attention mechanism in playing MOBA games, we model the control problem as an action generation process. 
We devise a deep latent alignment neural network model to learn the mechanism of action generation, and 
an adaptive selective attention samp-

\begin{table}[H]\setlength{\tabcolsep}{1.6pt}
\caption{Online command response peformance.}
\vskip -0.15in
\label{tab:command_response_online}
\begin{center}
\begin{small}
\begin{tabular}{lccccr}
\hline
Indices  &  Assemble  &  Retreat & Attack  \\
\hline
Ratio of command executed by no AI & 0.283 & 0.09 & 0.205 \\
Ratio of command executed by 1 AI  & 0.304 & 0.291 & 0.338 \\
Ratio of command executed by 2 AI  & 0.277 & 0.357 & 0.31 \\
Ratio of command executed by 3 AI  & 0.135 & 0.262 & 0.147 \\
\hline
Response rate of warrior hero  &  0.163  & 0.249  & 0.17 \\
Response rate of wizard hero  &  0.285  &  0.383  & 0.339   \\
Response rate of shooter hero   &  0.237  &  0.396  &  0.25   \\
Response rate of assistant hero  &  \textbf{0.46}  &  \textbf{0.561}  &  \textbf{0.519}     \\
Response rate of jungle hero  &  0.119  &  0.202  &   0.121  \\
\hline
\end{tabular}
\end{small}
\end{center}
\vskip -0.15in
\end{table}

\begin{table}[H]\setlength{\tabcolsep}{1.6pt}
\caption{Online command execution peformance.}
\vskip -0.15in
\label{tab:command_execute_online}
\begin{center}
\begin{small}
\begin{tabular}{lccccr}
\hline
 & Indices &  Assemble  &  Retreat & Attack  \\
\hline
\multicolumn{2}{c}{Number of commands executed per game}  & 1.844 & 1.605 &  1.839 \\
\hline
Warrior hero & Success rate  &  0.513   & 0.673  & 0.566 \\
& Abnormal rate  &   \textbf{0.045}    &      0        &   0.048   \\
\hline
Wizard hero & Success rate  &  0.613  &  0.781 & 0.541   \\
 & Abnormal rate  &  0.049  &   0    &   0.061   \\
\hline
Shooter hero & Success rate  &  0.519  &  0.804  &  0.513   \\
 & Abnormal rate  &  0.046   & 0.0005  &  0.065  \\
\hline
Assistant hero & Success rate  &  \textbf{0.673}  &  0.685  &  \textbf{0.605}     \\
& Abnormal rate  &  0.047    &   0    &    0.064  \\
\hline
Jungle hero &  Success rate  &  0.44  &  \textbf{0.801}  &   0.56  \\
& Abnormal rate  &   0.061     &    0    &    \textbf{0.038}  \\
\hline
\end{tabular}
\end{small}
\end{center}
\vskip -0.15in
\end{table}

\noindent ling algorithm serving as mechanism of control. 
We design deterministic and stochastic attention implementations of the core latent alignment model. 
Both simulated and online experiments illustrate that deterministic and stochastic agents are able to execute human's command well. 
Moreover, we show that for the first time, agent with stochastic action generation can play MOBA games well. 

\newpage
\normalsize
\bibliographystyle{aaai}
\bibliography{ref}

\end{document}